# Semantic Shifts of Psychological Concepts in Scientific and Popular Media Discourse: A Distributional Semantics Analysis of Russian-Language Corpora

**Orlova Anastasia,**
HSE University
asorlova@hse.ru

## Abstract

This article examines semantic shifts in psychological concepts across scientific and popular media discourse using methods of distributional semantics applied to Russian-language corpora. Two corpora were compiled: a scientific corpus of approximately 300 research articles from the journals Psychology. Journal of the Higher School of Economics and Vestnik of Saint Petersburg University. Psychology (767,543 tokens) and a popular science corpus consisting of texts from the online psychology platforms «Ясно» (Yasno) and «Чистые когниции» (Chistye kogntsii) (1,199,150 tokens). After preprocessing (OCR recognition, lemmatization, removal of stop words and non-informative characters), the corpora were analyzed through frequency analysis, clustering, and the identification of semantic associations. The results reveal significant differences in vocabulary and conceptual framing between the two discourse types: scientific texts emphasize methodological and clinical terminology, while popular science materials foreground everyday experience and therapeutic practice. A comparison of semantic associations for key concepts such as burnout and depression shows that scientific discourse links these terms to psychological resources, symptomatology, and diagnostic constructs, whereas popular science discourse frames them through personal narratives, emotions, and everyday situations. These findings demonstrate a clear shift from precise professional terminology toward more generalized and experiential meanings in popular media discourse and confirm the effectiveness of distributional semantics methods for identifying semantic transformations of psychological concepts across different communicative contexts.

As part of the study, a text corpus was compiled, comprising two types of sources: scientific articles and popular science publications on the topic of psychology. The goal of data collection was to identify semantic shifts in the use of psychological terms in academic and popular science discourse.

The first corpus consists of research articles in psychology published in two Russian scientific journals: "Psychology. Journal of the Higher School of Economics" indexed in Scopus and Web of Science, and "Vestnik of Saint Petersburg University. Psychology" (767,543 tokens). The sample spans a five-year period and includes approximately 300 articles. All texts were obtained in PDF format, recognized using optical character recognition (OCR) tools, and then converted to plain text for subsequent processing and analysis.

The second corpus includes materials from online media platforms dedicated to psychology: «Ясно» (Yasno) and «Чистые когниции» (Chistye kogntsii) (1,199,150 tokens). These resources are commercial internet platforms aimed at promoting psychological services and popularizing psychological knowledge. Both projects publish short popular science texts on psychology, which often become the subject of critical discussions in comments from the professional community. The publication data was exported from the platforms' web pages in HTML format and then converted to plain text for further processing.

Comparing the vocabulary of two corpora, whose topics touch on the same subject matter, but in different languages and for different audiences, allows us to trace differences in the semantic shifts of psychological concepts in the context of distributional semantics.

I would like to note that the automatic extraction of named entities from texts published in Telegram channels resulted in the disappearance of a number of concepts and terms, which is likely due to the syntax of short and scientific and educational texts. Therefore, a decision was made to preserve the terms and concepts and avoid removing named entities. Both corpora were pre-processed: stop words were removed, lemmatization was performed, word breaks were eliminated, and non-informative characters (punctuation marks, special characters) were remove

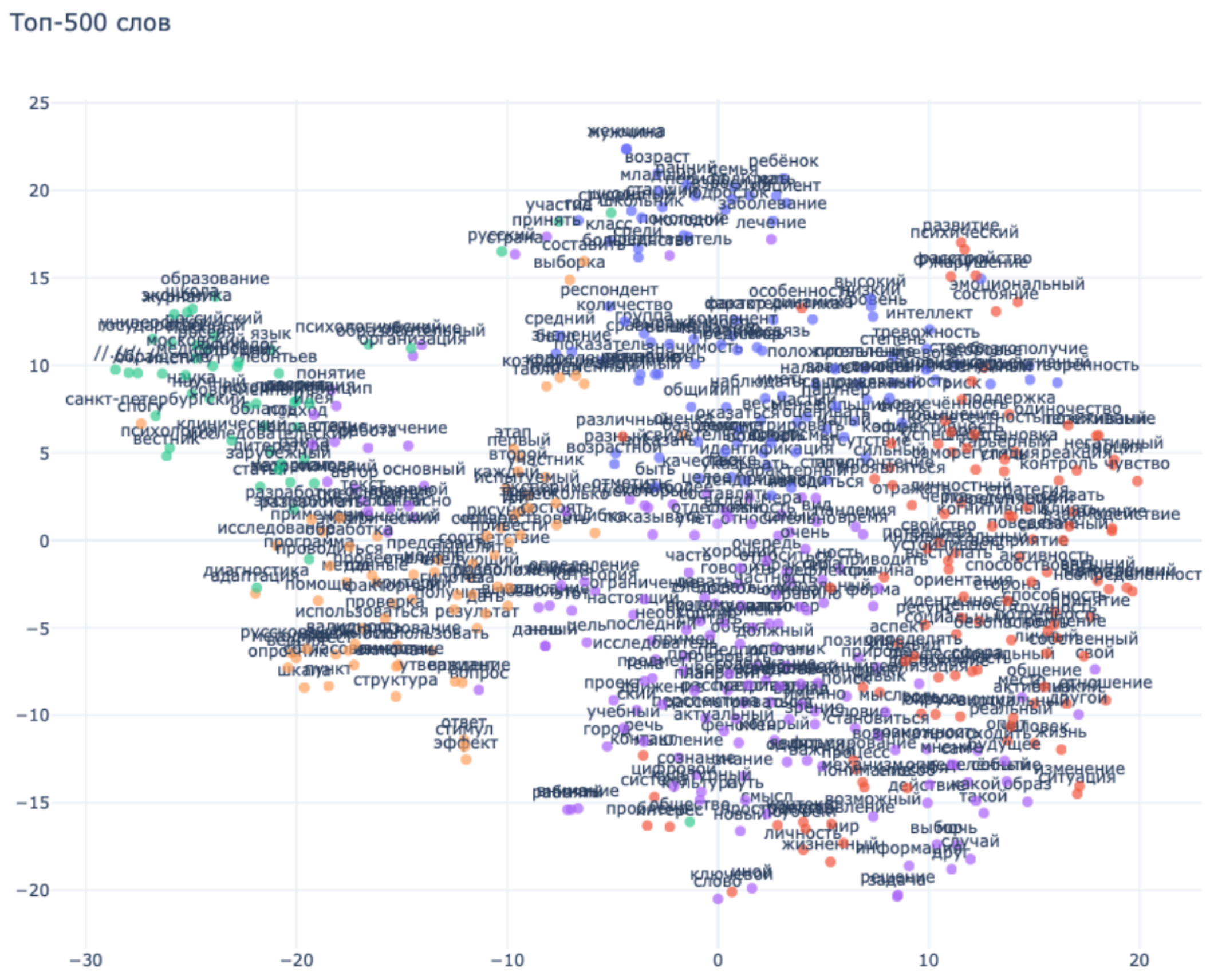

*Figure 1. Top 500 most frequent words in the scientific corpus*

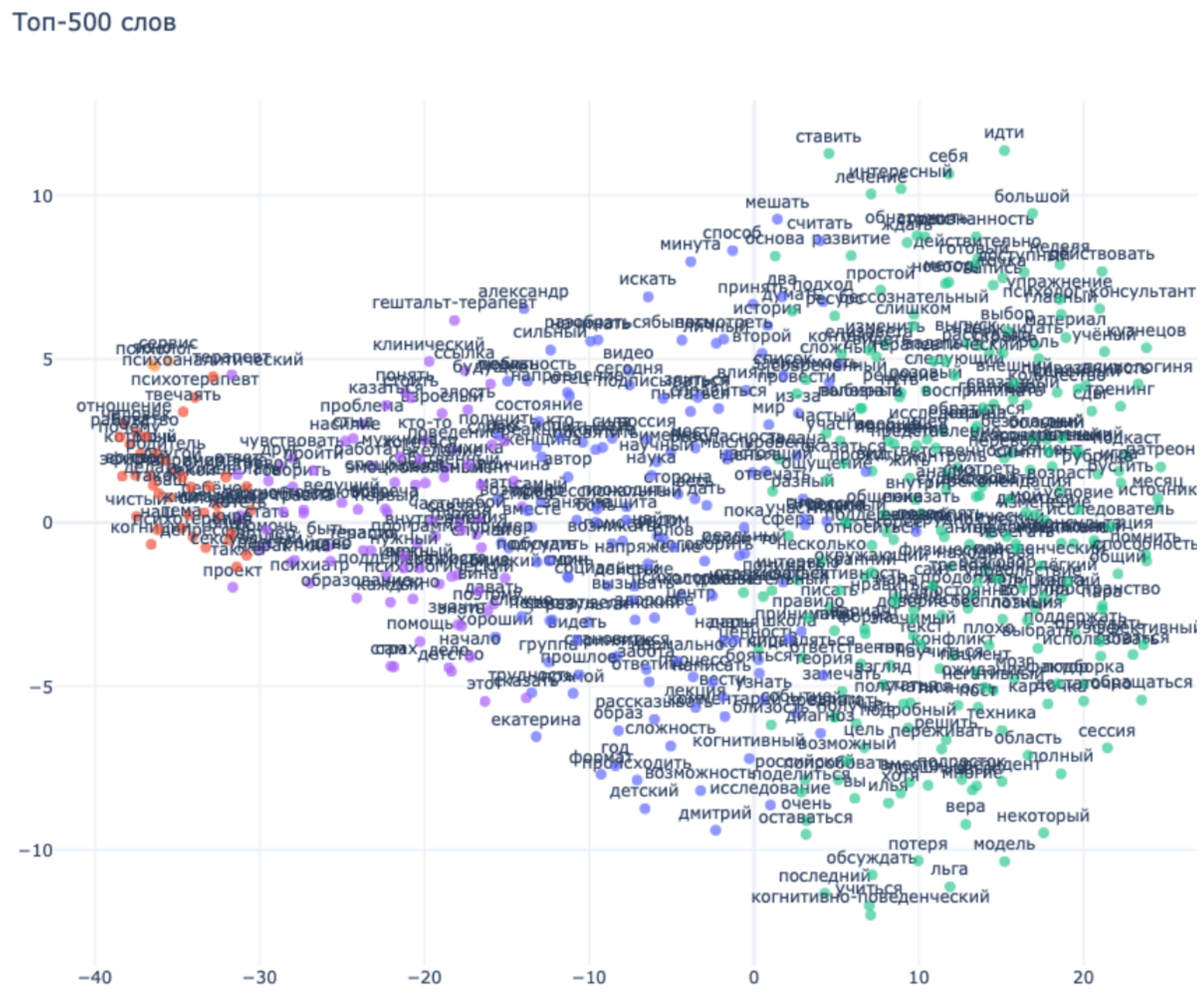

*Figure 2. Top 500 most frequent words in the popular science corpus*

Already at this stage, one can notice the differences in vocabulary in the two presented corpora, which cover the same field. For example, in the scientific context, the words «выборка» (sample), «этап» (stage), «возраст» (age), «шкала» (scale), and «структура» (structure) are prominent, referring to statistical studies, while in the popular science context, there is no emphasis on scientific research. However, the words «идти» (go), «сервис» (service), «сессия» (session), «конусльтант»(consultant), and «гештальт-терапевт» (gestalt therapist) are present, as are the prominent names (probably therapists and authors of the channel), which alludes not to a theoretical understanding of the scientific field, but to the importance of the practical side of the issue (therapy).

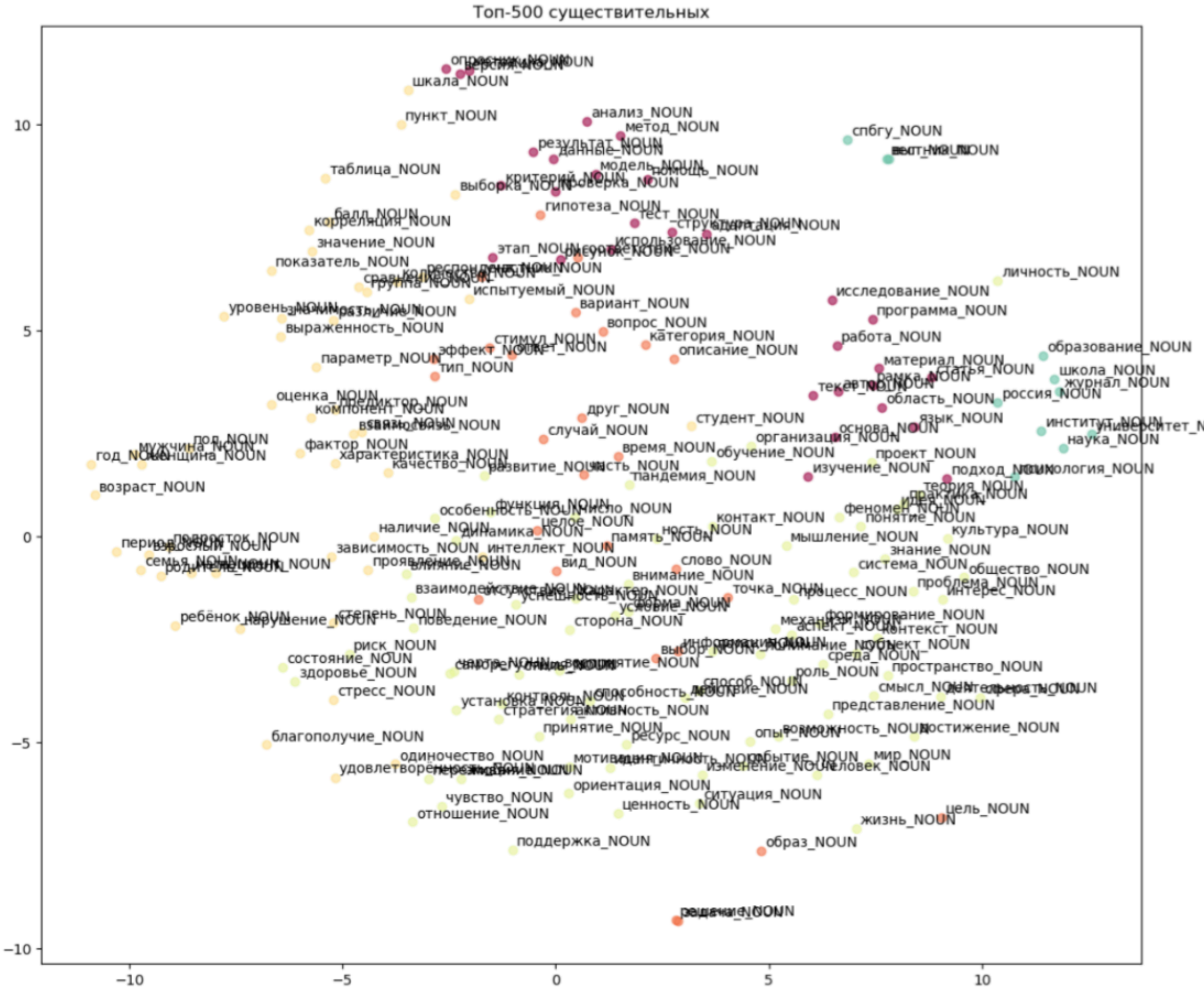

*Figure 3. Top 500 most frequent nouns in the scientific corpus*

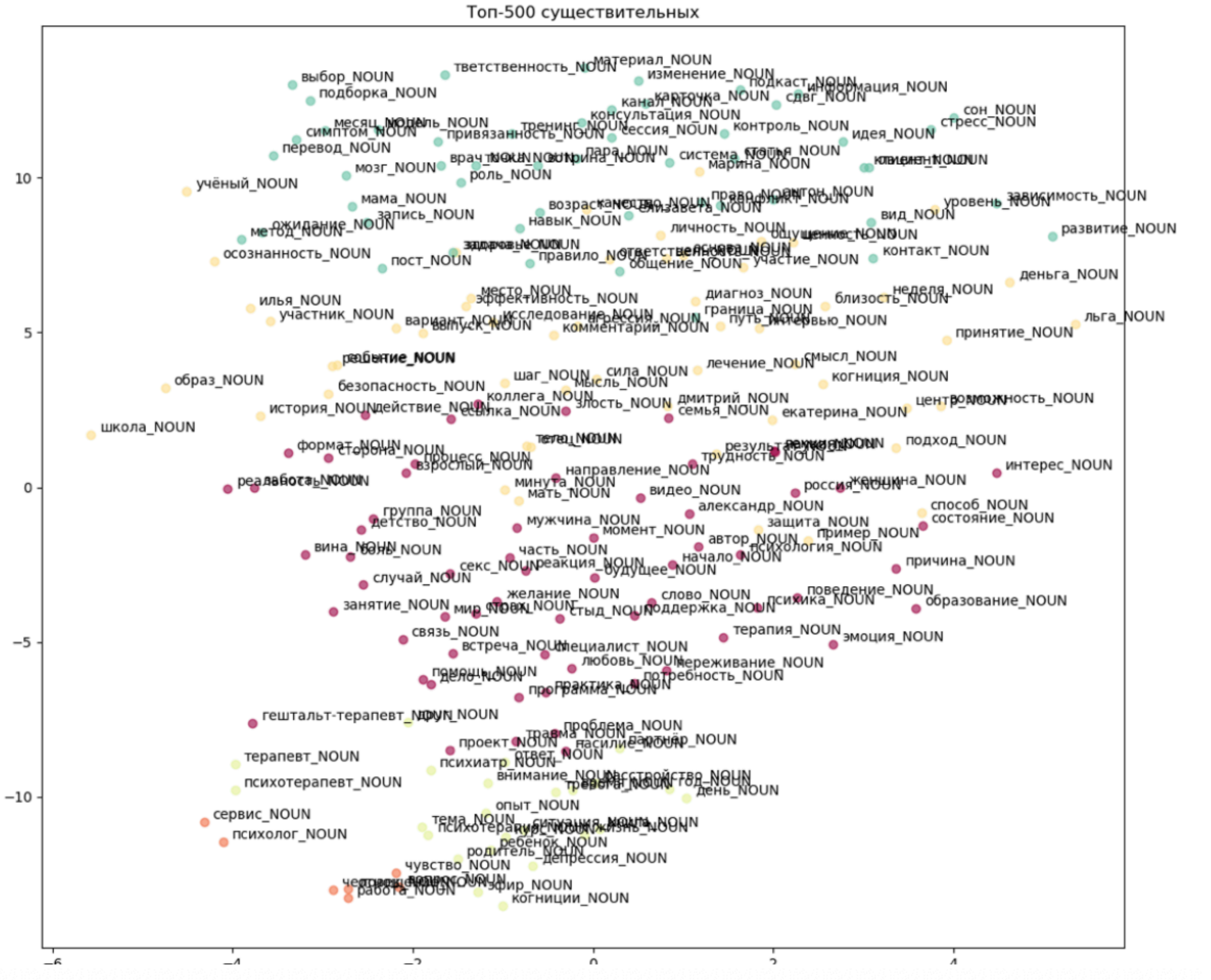

*Figure 4. Top 500 most frequent nouns in the popular science corpus*

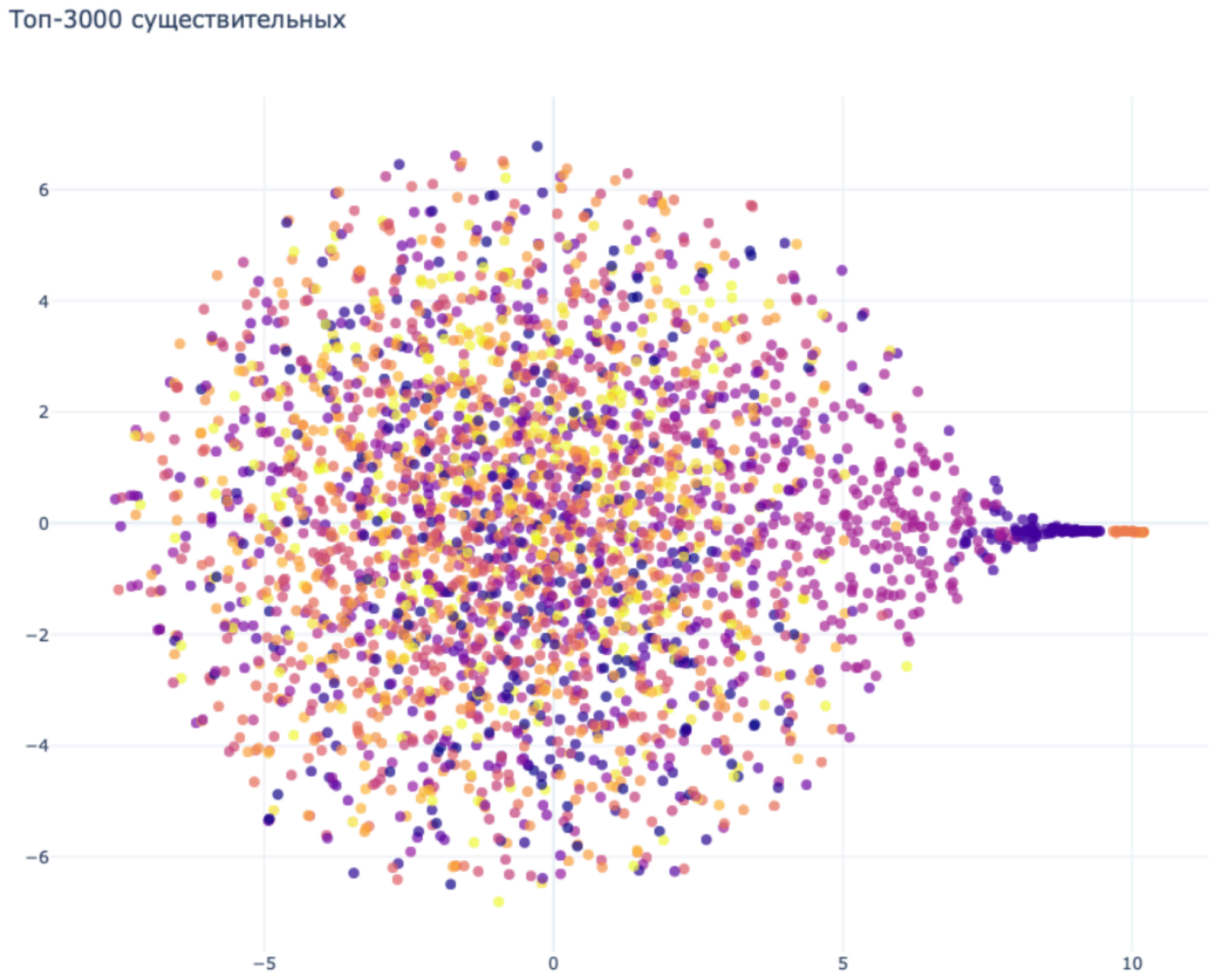

*Figure 5. Clustering of texts in the Telegram channel corpus*

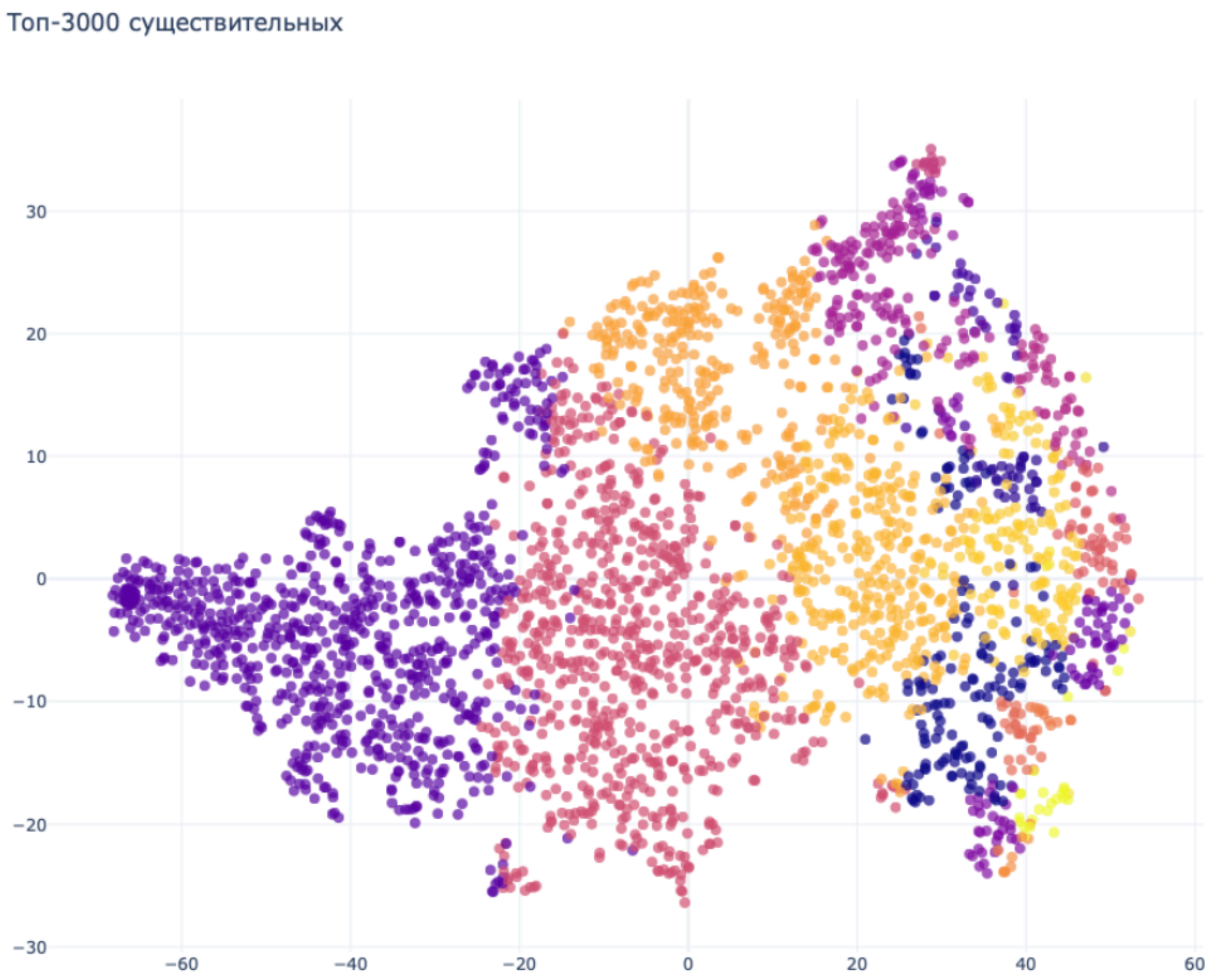

*Figure 6. Clustering of texts in the scientific corpus*

Clustering shows disunity for the corpus of texts from the Telegram community (with the exception of a small "outlier" on the right), which may be due to the length of the text, as well as the style and format of presentation, while the corpus of scientific texts is well clustered into groups that can be interpreted.

**Semantic associates.**

Semantic associates for the word **«выгорание» (burnout)** in a scientific context are: «родительский» (parental), «утомление» (fatigue), «повышение» (increase), «жизнестойкость» (resilience), «пользовательский» (user), «автономность» (autonomy), «самооценка» (self-esteem), «стрессоустойчивость» (stress resistance), «самоуважение» (self-respect), «развитость» (development), «повыситься» (increase), «фрустрированность» (frustration), «посттравматический» (post-traumatic), «дисфункциональный» (dysfunctional),

«травмированный» (traumatized), «маркер» (marker), «интернальность» (internality), «пограничным» (iborderline), «клинический» (clinical), «сформированность» (maturity), «дистресс» (distress).

Semantic associations for the word **«выгорание» (burnout)** in popular science contexts include: «эфир» (ether), «работа» (work), «наш» (our), «человек» (person), «жизнь» (life), «отношение» (attitude), «другой» (differend), «почему» (why), «психотерапия» (psychotherapy), «такой» (such), «вопрос» (question), «например» (for example), «ситуация» (situation), «который» (which), «мочь» (can), «родитель» (parent), «чистый» (pure), «чувство» (feeling), «стать» (become).

Let's turn to the definition of the word from Russian Wikipedia (when accessing this site, we rely on the principle of "wisdom of crowds"). Burnout syndrome is a concept introduced into psychology by the American psychiatrist Herbert Freudenberger in 1974, characterized by growing emotional exhaustion. It can lead to personality changes in social interactions (even to the development of profound cognitive distortions). The term "burnout syndrome" itself was coined by K. Maslach in accordance with the three-component model of "burnout." "Burnout" was defined as a syndrome and included three components: «эмоциональное истозение» (emotional exhaustion), «деперсонализация» (depersonalization), and a «редукция личных достижений» (reduction in personal achievements) — a diminished sense of self-worth.

Semantic associations for the word **«депрессия» (depression)** in a scientific context include: «тревога» (anxiety), «дистресс» (distress), «алекситимия» (alexithymia), «астенический» (asthenic), «одинокость» (loneliness), «интенсивность» (intensity), «подавленность» (depression), «астения» (asthenia), «ПТСР» (PTSD), «гнев» (anger), «послеродовой» (postpartum), «тревожность» (anxiety), «обсессивно-компульсивный» (obsessive-compulsive), «эгоизм» (egoism), «эго-благополучие» (ego well-being), «посттравматический» (posttraumatic), «фрустрация» (frustration), «повышенный» (increased), «депрессивный» (depressive), «симптом» (symptom).

Semantic associations for the word **«депрессия» (depression)** in popular science contexts include: почему (why), эфир (ether), работа (work), отношение (attitude), свой (one's own), человек (person), вопрос (question), такой (such), чувство (feeling), мочь (can), который (which), психотерапия (psychotherapy), родитель (parent), другой (other), книга (book), делать (do), ваш (your), какой (what), новый (new), например (for example).

Let's turn again to the Russian Wikipedia definition. Depression (from the Latin deprimo "to press (down), suppress") is a mental disorder whose main symptoms are a depressed— depressed, depressed, melancholy, anxious, fearful, or indifferent —mood and a decrease or loss of the ability to experience pleasure (anhedonia). Typically, some of the following symptoms are also present: low self-esteem, loss of interest in life and in usual activities, inappropriate feelings of guilt, pessimism, impaired concentration, fatigue or lack of energy, sleep and appetite disturbances, and suicidal tendencies. Severe forms of depression are characterized by the so-called "depressive triad": depressed mood, slowed thinking, and motor retardation.

Thus, an analysis of the semantic associations of the words «выгорание» (burnout) and «депрессия» (depression) in scientific and popular science contexts shows how significantly approaches to these concepts differ depending on the purpose audience.

In a scientific context, the word «выгорание» (burnout) is associated with terms related to psychological characteristics and resources: fatigue, stress resilience, self-esteem, autonomy, resilience, and maturity. There are also markers indicating a potential clinical picture: frustration, post-traumatic, dysfunctional, traumatized, and clinical. This emphasizes that burnout in science is perceived not simply as fatigue, but as a multi-level psycho-emotional and personal phenomenon, ncluding internal assessments and the ability to adapt.

At the same time, «депрессия» (depression) in the scientific field is surrounded by more pronounced medical and clinical concepts: «тревога» (anxiety), «депрессия»(depression), «фрустрация» (frustration), «астения» (asthenia), «ПТСР» (PTSD), «обсессивно-компульсивный» (obsessive-compulsive), and «послеродовой»(postpartum). Here, the focus shifts to symptomatology and diagnostic parameters, addressing underlying emotional states («гнев» (anger), «тревожность» (anxiety), as well as personal constructs («эгоизм» (egoism), «эго-благополучие» (ego-well-being).

Thus, «выгорание» (burnout) is described as a reaction to external stress and depletion of internal resources, while depression is presented as a systemic, endogenous, or reactive disturbance of the psychoemotional state.

In popular science, both burnout and depression show a sharp shift toward everyday, mundane, and personal concepts. Associated terms include words like «работа» (work), «человек» (person), «чувство» (feeling), «родитель» (parent), «ситуация»(situation), «почему» (why), «например» (for example), «психотерапия»(psychotherapy). These words indicate a shift into the realm of personal experience. There is no mention of clinical diagnoses or theoretical constructs. From popular science contexts, we can learn nothing about the semantic component of the concepts discussed.

If we turn to the Russian Wikipedia definition of «выгорание» (burnout), it confirms the scientific view: burnout is a syndrome comprising three key components — «эмоциональное истощение» (emotional exhaustion), «деперсонализация»(depersonalization), and a «редукция личных достижений» (reduction in personal achievemen)t. All these elements are reflected in scientific associations: for example, «фрустрация» (frustration) indicates emotional exhaustion, «клинический» (clinical) indicates a diagnostic threshold, and «самооценка» (self-esteem) indicates a reduction in achievement.

For depression, Russian Wikipedia lists key features — «подавленное настроение» (depressed mood), «ангедония» (anhedonia), «низкая самооценка» (low self-esteem), «пессимизм» (pessimism), as well as physiological and cognitive symptoms. Almost all of them are reflected in scientific associations: «тревога» (anxiety), «фрустрация» (frustration), «ПТСР» (PTSD), «депрессия» (depression), as well as specific forms («послеродовая» (postpartum), «обсессивно-компульсивный» (obsessive-compulsive). This indicates a high consistency between the definition and the actual semantic perception of the term in the scientific community.

Overall, several key conclusions can be drawn. Scientific discourse around «выгорание» (burnout) focuses on the individual and adaptation to stress, while «депрессия» (depression) is viewed as a disorder with pronounced clinical symptomatology. In popular science, these distinctions are blurred — both concepts become elements of personal stories, emotional experiences, and everyday narratives. This demonstrates the gap between the precision of professional language and the vagueness and promotional style of popular science.

A similar picture is observed for the popular definitions of «осознанный» (mindful) and «посттравматический» (post-traumatic).

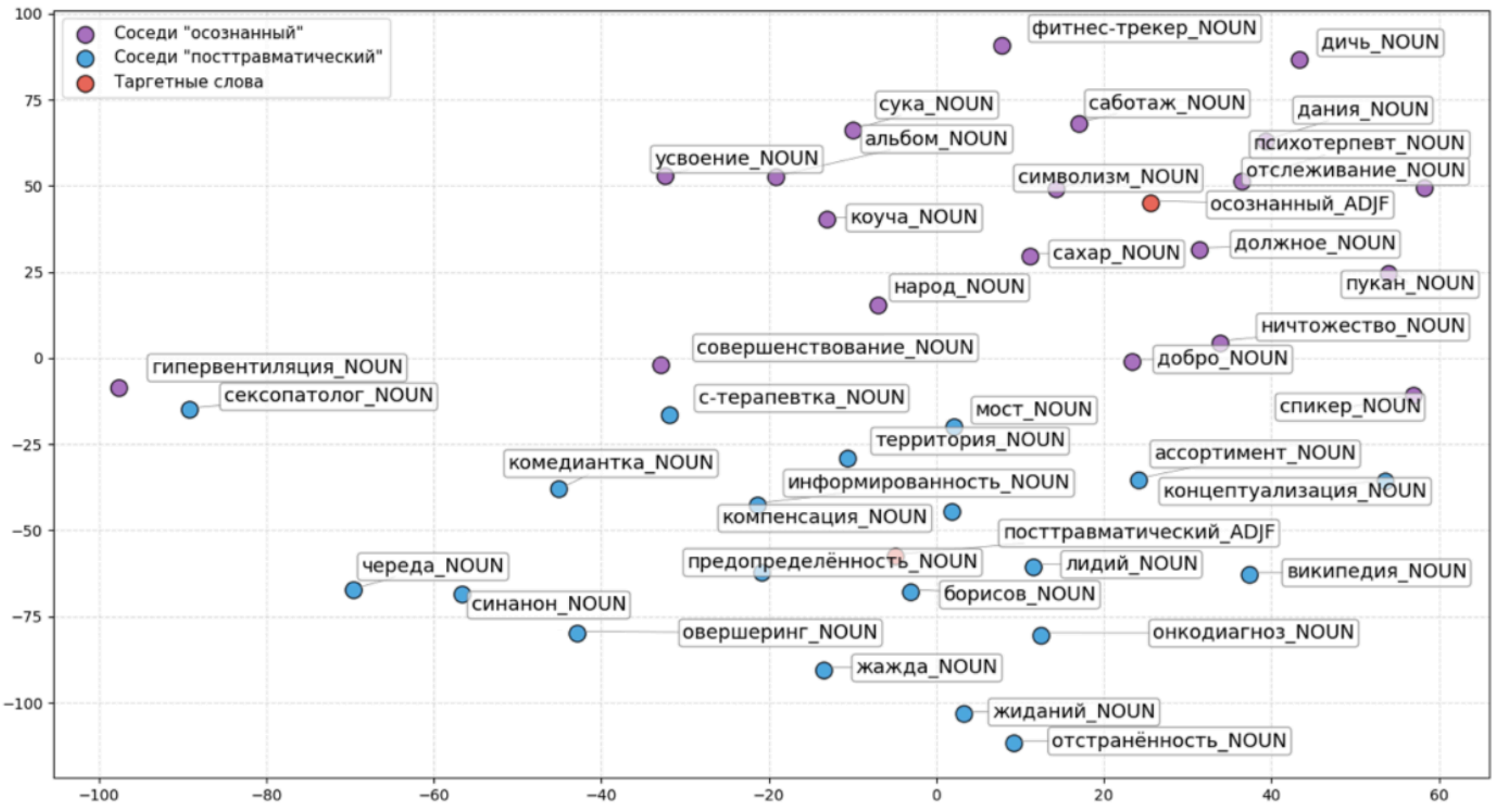

*Figure 7. Semantic associations of the words mindful and post-traumatic in the scientific corpus*

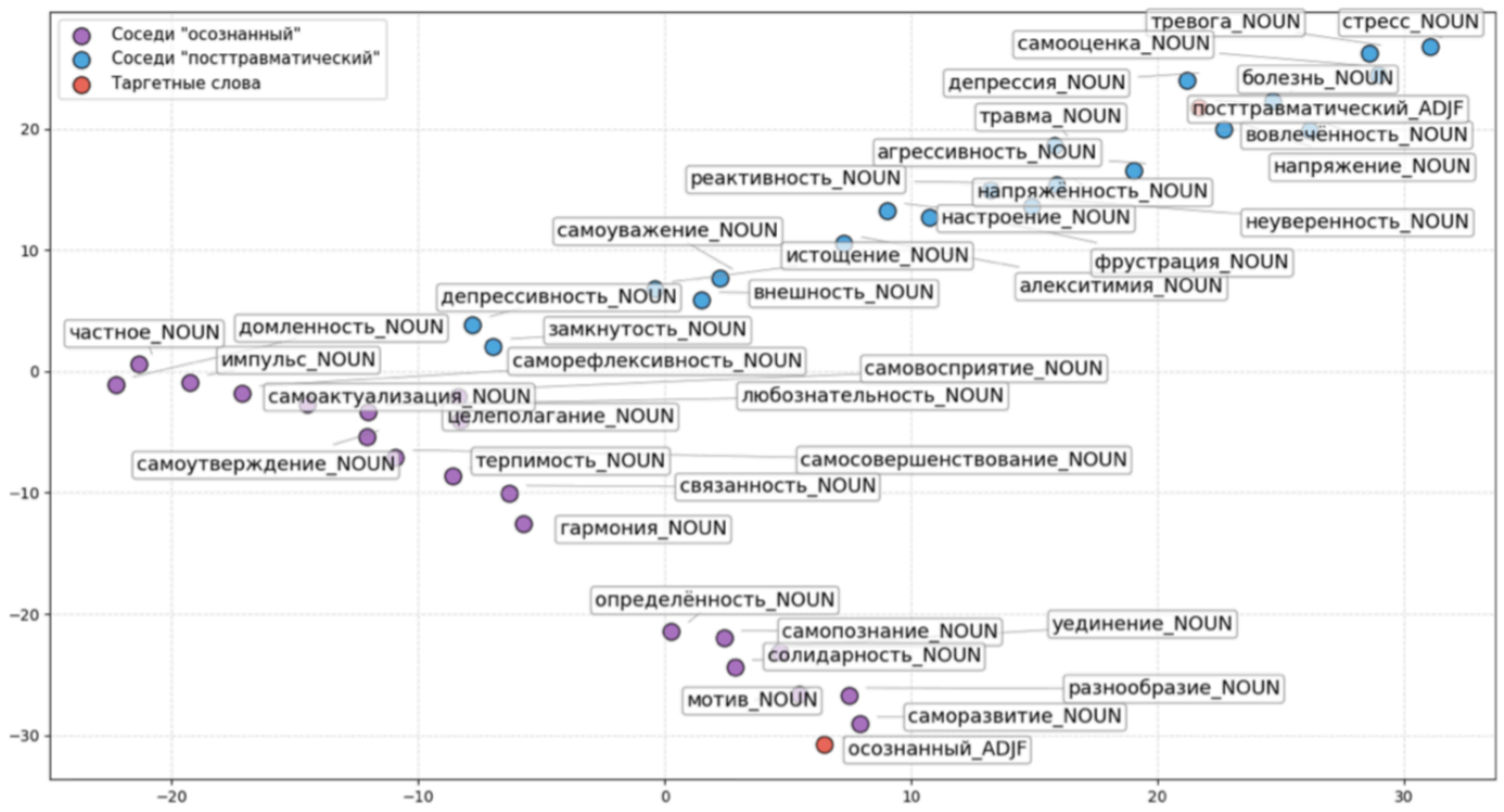

*Figure 8. Semantic neighbors of the words mindful and post-traumatic in the scientific and popular science corpora*

Thus, methods of distributional semantics have proven effective or tracking semantic shifts and semantic transformations of concepts traditionally found in scientific contexts.

**Acknowledgements**

I would like to express my sincere gratitude to Olga Alieva for her inspiration and for the knowledge she has shared with me over time. I would also like to thank Alexander Klimov, within whose course this work was developed. I further thank Boris Orekhov for his valuable comments on this work. Any remaining errors or shortcomings are solely my responsibility.

**References**

Boris Orekhov. Individual semantics of L. N. Tolstoy in the context of vector models. 14(4):119–129, 2023. ISSN 2782-5450. doi:10.18721/JHSS.14409. URL https://human.spbstu.ru/en/article/2023.54.09/.

Liétard, B., Keller, M., & Denis, P. (2023). A Tale of Two Laws of Semantic Change: Predicting Synonym Changes with Distributional Semantic Models. Proceedings of the 12th Joint Conference on Lexical and Computational Semantics (SEM 2023), 338–352.

Rodda, M. A., Senaldi, M. S., & Lenci, A. (2017). Panta rei: Tracking semantic change with distributional semantics in Ancient Greek. IJCoL. Italian Journal of Computational Linguistics, 3(3-1), 11-24.

Rodina, J., & Kutuzov, A. (2020, December). RuSemShift: a dataset of historical lexical semantic change in Russian. In Proceedings of the 28th International Conference on Computational Linguistics (pp. 1037-1047).

Schlechtweg, D., Hätty, A., Del Tredici, M., & Im Walde, S. S. (2019, July). A wind of change: Detecting and evaluating lexical semantic change across times and domains. In Proceedings of the 57th annual meeting of the association for computational linguistics (pp. 732-746).

Tang, X., Zhou, Y., Aida, T., Sen, P., & Bollegala, D. (2023, December). Can word sense distribution detect semantic changes of words?. In Findings of the Association for Computational Linguistics: EMNLP 2023 (pp. 3575-3590).